\documentclass[sw,crcready]{iosart2x}

\usepackage[utf8]{inputenc}
\usepackage{xurl}
\usepackage{tabularx}
\usepackage[T1]{fontenc}
\usepackage{rotating}
\usepackage{adjustbox}
\usepackage{makecell}
\usepackage{color}
\usepackage{hyperref}
\usepackage{cleveref}
\usepackage{url}
\usepackage{makecell}
\usepackage{booktabs}
\usepackage{listings}
\usepackage{comment}

\pubyear{0000}
\volume{0}
\firstpage{1}
\lastpage{1}

\begin{document}

\begin{frontmatter}

\title{Building a Knowledge Graph of Distributed Ledger Technologies}
\runtitle{DLT Knowledge Graph}

\begin{aug}
\author{\inits{L.}\fnms{Lukas} \snm{König}\ead[label=e1]{lukas.koenig@fhstp.ac.at}%
\thanks{Corresponding author. \printead{e1}.}}
\author{\inits{S.}\fnms{Sebastian} \snm{Neumaier}\ead[label=e2]{sebastian.neumaier@fhstp.ac.at}}
\address{\orgname{St. Pölten University of Applied Sciences},
\cny{Austria}\printead[presep={\\}]{e1}}
\end{aug}

\maketitle              


\begin{abstract}
Distributed ledger systems have become more prominent and successful in recent years, with a focus on blockchains and cryptocurrency. 
This has led to various misunderstandings about both the technology itself and its capabilities, as in many cases blockchain and cryptocurrency is used synonymously and other applications are often overlooked.
Therefore, as a whole, the view of distributed ledger technology beyond blockchains and cryptocurrencies is very limited. Existing vocabularies and ontologies often focus on single aspects of the technology, or in some cases even just on one product. 
This potentially leads to other types of distributed ledgers and their possible use cases being neglected.
In this paper, we present a knowledge graph and an ontology for distributed ledger technologies, which includes security considerations to model aspects such as threats and vulnerabilities, application domains, as well as relevant standards and regulations. Such a knowledge graph improves the overall understanding of distributed ledgers, reveals their strengths, and supports the work of security personnel, i.e. analysts and system architects. 
We discuss potential uses and follow semantic web best practices to evaluate and publish the ontology and knowledge graph.

\vspace{5pt}
\noindent \textbf{URI:} \url{https://w3id.org/DLTOntology}\\
\textbf{DOI:} \href{https://doi.org/10.5281/zenodo.6497619}{10.5281/zenodo.6497619}
\end{abstract}


\end{frontmatter}

\section{Introduction} \label{sec:intro}
While the success of blockchains and especially cryptocurrencies has helped to spread the word about distributed ledger technology (DLT), it has lead to misunderstandings about what the technology actually consists of and what it is capable of. In \cite{BCkleenex} this is highlighted very well with a comparison to facial tissues and the brand product Kleenex, which often get used interchangeably, even though the latter is just one implemented variant of the former. Such a misunderstanding of the technology can hinder its adoption. Each variant of distributed ledger technology operates in a slightly different way with regard to distinct requirements, which therefore means that there are also different usage scenarios and implementations that can be realized with distributed ledgers. 
Limiting the broad field of the technology to just one single variant of it leads to a distorted over-representation and use of that one variant, because the valid alternatives are simply not emphasized in the same way. There are observable attempts of reverting this trend by stating the technicalities and comparing different types of distributed ledger systems to one another, with new research and reviews stating up to 5 different types that are regularly seen as equally valid implementations of a distributed ledger system \cite{el2018review,panwar2020distributed}.\footnote{\url{https://i.pinimg.com/originals/9d/d1/1b/9dd11bbaf5f025f3bd345e6816a4fc16.png}} \footnote{\url{https://medium.com/@support_61820/different-types-of-dlts-and-how-they-work-cfd4eb218431}}

One way to unify the attempts of providing clarity on distributed ledgers is to use a knowledge graph to capture different technical capabilities and the wider ecosystem. In recent years, the concept of knowledge graphs has become increasingly popular, the idea being to use a graph-based data model to collect and convey knowledge about the real world \cite{10.1145/3447772}. To define the semantics of the terms used in the graph, an ontology, i.e. a formal representation of the relationships and classes in the graph, is used \cite{ontology101}.
As we point out in detail in \Cref{sec:relatedWork}, an all-inclusive approach towards a distributed ledger ontology is currently still lacking. Existing ontologies and vocabularies typically focus on a limited sub-group of distributed ledger systems, or build an ontology for a single distributed ledger product or application. Blockchains and blockchain-based systems are often the focus of such works, which leaves a large part of distributed ledgers unaccounted for. 

This work therefore establishes a modular ontology, which encompasses and relates the following concepts and areas: Distributed ledger technology with a broader perspective than just a limited focus on blockchains, broader security considerations and technical implications of such systems, as well as organizational boundaries and real world applicability. 
Additionally, this ontology and the respective knowledge graph aims to clarify the distinctions between different types and systems of distributed ledgers, as it includes a broader spectrum than merely the view on blockchain as the paragon of the entire technology.

The model and subsequent knowledge graph created in this work can be used in a multitude of ways: Besides the already mentioned perspective of the entire ecosystem, it also allows for a more distinct observation of the included subjects. For instance, it can be used for threat analysis and risk assessments by identifying attacks and vulnerabilities of a system. In particular, security analysts will benefit from such a model, as it enables them to store and retrieve critical information about systems they might use, and to highlight further measures where they are needed to secure a system.

The main contributions of this work can be summarized as follows:
\begin{itemize}
    \item a survey of literature and information on distributed ledger technologies, use-cases, fields of application, standardizations and legal issues;
    \item a set of competency questions that cover the main aspects of the collected information;
    \item an RDFS-based ontology that allows to model the surveyed aspects;
    \item a knowledge graph of relevant entities and relations, extracted from the literature;
    \item an evaluation of the model via SPARQL queries; an online interface to execute the queries.
\end{itemize}

The remainder of this paper is structured as follows: \Cref{sec:DLTbackground} provides an overview about distributed ledger technologies, as well as a short introduction to Ontologies, Knowledge Graphs and their terminologies. \Cref{sec:method} details the methods, i.e., the information collection, the competency questions and evaluation of the ontology. \Cref{sec:Ontology} describes the ontology in detail, and explains the elements and contents of the knowledge graph. In \Cref{sec:relatedWork} we discuss related work and conclude in \Cref{sec:conclusion}.


\section{Background} \label{sec:DLTbackground}



\subsection{DLT Data Structures} \label{subsec:dataStructures}
\paragraph{Blockchain.} \label{subsec:blockchain}
The most well-known example of a distributed ledger technology is a blockchain system. Especially the specific blockchain application Bitcoin sparked a hype around cryptocurrencies that lead to an often misunderstood terminology where Bitcoin and Blockchain are used interchangeably \cite{carson2018blockchain,deshpande2017distributed}, which is not true however. There are many more blockchain applications; cryptocurrencies are merely one aspect of the broader picture. Other examples of blockchain technologies are Ethereum and its renewed introduction of smart contracts, which have become increasingly popular over the last years, as well as the Hyperledger project by the Linux Foundation and IBM \cite{shrivas2020disruptive}.\\
At its technological core, a blockchain is a data structure that (cryptographically) links blocks of data into a chain of blocks. Blockchains are usually further separated into either public, private or federated blockchains, according to the degree of access that is implemented where a public blockchain is accessible to anyone and a private blockchain is restricted and usually used internally \cite{el2018review,rawat2019blockchain}.
\paragraph{Directed Acyclic Graph.} \label{subsec:DAG}
Directed Acyclic Graphs (DAG) as a data structure used for distributed ledger systems started to emerge with the applications NXT, and IOTA with its \emph{Tangle}, which is especially focused on IoT, micro transactions and devices with lower computational power \cite{el2018review,panwar2020distributed}. Instead of having data blocks hashed into one single line, DAGs are used where connections between transactions in the graph show where a valid transaction is present. Since each transaction is processed individually and does not require the formation of blocks as with a regular blockchain, there is technically no transaction limit for DAGs and therefore they are very scaleable.
\paragraph{Hashgraph.} \label{subsec:Hashgraph}
The most prominent product for Hashgraphs is \emph{Hedera Hashgraph} as a next generation DLT that focuses on fairness and transaction speed with regard to the order of events as they unfold. It is very efficient to use as it uses virtual voting for reaching consensus and information about transactions are shared with a gossip protocol. Technically, the data structure used for Hashgraphs is similar to a DAG, which is bound to the sequence of time and the nodes participating in the network. By that they fundamentally differentiate themselves from regular DAGs \cite{el2018review,panwar2020distributed}.
\paragraph{Holochain.} \label{subsec:Holochain}
Contrary to the previous types of DLTs, the major change with \emph{Holochain} is the shift from a focus on data itself towards an agent-centric structure which some consider the purest form of distributed ledger, as every node virtually is responsible for its own ledger. Validation is done by the introduction of a ruleset to the Holochain network. This ruleset is referred to as DNA and can be used to verify a node and to spot malicious actors. The changes introduced with Holochain make mining obsolete, which makes it a very energy efficient alternative \cite{panwar2020distributed}.


\subsection{Security Challenges, Opportunities and Use Cases for Blockchains \& Distributed Ledger Systems} \label{sec:dltthreats}
As with regular ICT systems, distributed ledger technologies are not immune to threats. While the introduction of distributed ledgers by itself can mitigate certain risks already, it also introduces new dangers that need to be considered. Especially since there are a number of myths revolving around the technology that make it seem almost invincible \cite{carson2018blockchain,shrivas2020disruptive}.
\paragraph{Organizational Challenges.}
One of the challenges of using blockchains for supply chains or as a mean of ensuring transparency and auditability is that if manual input is required, the people involved are still prone to errors and false input \cite{jabbar2020blockchain}. Additionally, the knowledge about and familiarity with distributed ledgers is still limited generally speaking. This lack of knowledge also hampers the willingness to become an early adopter of the technology \cite{jabbar2020blockchain,deshpande2017distributed,carson2018blockchain}. Another challenge would be a lack of standardization for distributed ledgers or the integration into existing standardization frameworks or questions about governance, legal requirements, and regulations, especially for cyber security \cite{fi12120222,gramoli2018blockchain}. This challenge is only aggravated by the fact that there is no unified terminology or vocabulary as well \cite{de2017towards}.

\paragraph{Technical Challenges.}
So far there has been intensive research on the vulnerabilities and attacks on distributed ledger systems. A major part of these focuses on blockchain systems where \emph{Bitcoin} and \emph{Ethereum} are often times the center of attention. Examples for blockchain attacks are 51\%-Attacks, Fork Attacks, and Selfish Mining \cite{jisis20-10-3-06}. Additionally, blockchains are often affected by scalability issues and low transaction rates due to mining and consensus mechanisms \cite{chen2020survey,saad2020exploring}. For other forms of distributed ledger, like a DAG, there is a strong focus on IoT and IIoT \cite{sengupta2020comprehensive}, where security features that are present in a blockchain structure are sacrificed for an increase in performance.

\paragraph{Opportunities and Use Cases \label{sec:dltusecases}}
There are plenty of other fields of use for distributed ledger systems outside of the world of cryptocurrencies. Since that is where the hype started, the introduction of other financial applications that use the technology came to no surprise. This includes not only payment services but also the financial infrastructure and international finance for example \cite{rawat2019blockchain,schlegel2018blockchain}.
However, distributed ledgers are also already extensively used in supply chains around the globe as well. Instances of that would be shipping and logistics, ethical supply chains, or food safety \cite{jabbar2020blockchain,carson2018blockchain}. 
Another opportunity for distributed ledgers is their use for government services, be it for voting, record-keeping, transparency or combating fraudulent activities \cite{rawat2019blockchain,schlegel2018blockchain}.
Other recommended and proposed fields for the use of distributed ledger technology are for example healthcare, identity management, the internet of things, notary and insurance services, and (intellectual) property ownership \cite{kolanmedical,carson2018blockchain}.



\section{Methodology\label{sec:method}}

Methodologically, our ontology creation process follows the recommendations by Noy and McGuiness \cite{ontology101}. The modelling and creation of the ontology followed an iterative process and was done using Protégé. The applied steps can be summarised as follows:

\begin{enumerate}
\item In the first step we determine the \textit{domain and scope} of the ontology. 
\item We gather literature about distributed ledger technologies, their components and existing standards, application domains, security vulnerabilities and threats (cf. Sec. \ref{ssec:informationcollection}).
\item We define a set of questions ``that a knowledge base based on the ontology should be able to answer.'' \cite{ontology101} We derive use cases and create \textit{competency questions} based on the collected information (cf. Section \ref{sec:evaluation}).
\item We create an ontology to express the collected information (cf. \Cref{sec:Ontology}).
\item We evaluate the ontology by providing SPARQL queries to answer the collected competency questions.
\item Finally, we instantiate the developed ontology with named entities and relations extracted from the literature (cf. \Cref{ssec:kgDetails}).
\end{enumerate}


\subsection{Ontology Scope\label{ssec:scope}}
The scope of the ontology is a model for a holistic, modular approach on the field of distributed ledger technology that does not focus on a single or a subset of technologies, but rather the wider distributed ledger ecosystem. The goal is to include factors which affect the implementation and operation of distributed ledgers, including existing standards and its usage across the industries.

As this ontology does not serve as an overview of one single product or technology stack, the technological composition of a system is merely a starting point for a broader observation that includes its strengths and weaknesses, as well as the general applicability of a system for specific use cases. It therefore serves as an application-neutral knowledge base for the broader field of distributed ledger technology.

However, to achieve this general applicability it was necessary to focus on the broader picture, which means that detailed technical implementations and processes like message flows and triggers, and involved accounts are out of scope. 


\subsection{Information Collection\label{ssec:informationcollection}}

The core concept of this ontology encompasses the three major areas \emph{technology}, \emph{business \& market use}, and \emph{legal \& standardization}.

In \Cref{sec:DLTbackground} we already discussed major structural differences between distributed ledger systems, including challenges and threats. The research in this mentioned section forms the basis for the technical part of the ontology, with additional information from further comparisons\footnote{\url{https://i.pinimg.com/originals/9d/d1/1b/9dd11bbaf5f025f3bd345e6816a4fc16.png}}. Additionally, technological differences between variants of blockchain/distributed ledger systems or smart contract platforms have been discussed and compared in \cite{dinh2018untangling,li2021services}, from where we drew additional input for the ontology.
Further information on the technological composition of distributed ledgers and possible technology stacks can be found in \cite{BCtechstack,teFoodWhite} and from technical reviews\footnote{\url{https://101blockchains.com/web-3-0-blockchain-technology-stack/}}.

For use cases and the real-world application of blockchains and distributed ledgers, we build upon existing assessments and research on the possibilities and likelihood of distributed ledgers being used as a tool in different business sectors, which can be found in e.g. \cite{rawat2019blockchain,carson2018blockchain}\footnote{\url{https://101blockchains.com/blockchain-digital-transformation/}}\footnote{\url{https://101blockchains.com/wp-content/uploads/2019/01/Applications-of-Distributed-Ledger-Technology-DLT.png}}. On top of that, these business sectors can further be split up into specific use cases, with a majority of these being mentioned in Section \ref{sec:dltusecases} as well.

When it comes to organizational standardization of distributed ledger technology, finished work that is entirely focused on it is scarce. Reports like \cite{deshpande2017distributed} offer a prediction and assessment of when and what to expect regarding standardization areas, others see it as a hurdle which is responsible for slow adoption rates of the technology, as mentioned in \ref{sec:dltthreats}. However, there are many standards in the works and several standardization organizations have set up focus groups or committees to create new normative standardization reference material specifically catering towards distributed ledger technology and blockchain \cite{fi12120222}.

Legal issues on the other hand are still mostly unresolved and require further considerations \cite{rodrigues2018law}. Especially when it comes to governance and (shared) responsibility or liability when using blockchains. The introduction of specified laws and a legal basis, in particular for governance and adoption of the technology, is seen as a crucial factor by the authors of \cite{werbach2018trust}. With ever rising numbers in cyber crime and based on the existing catalogues of laws on technology, it can be expected that distributed ledgers will receive their own set of laws in the future.

\subsection{Evaluation} \label{sec:evaluation}

The competency questions in Table \ref{tab:questions} are based on the collected information (see section \ref{ssec:informationcollection}) and grouped into three main categories:
\begin{itemize}
\item \textit{Technology and Security} involves questions regarding the components of a DLT system, as well as technical threats and vulnerabilities of systems and components. From this category we derive relations between vulnerabilities of the respective systems and components, and attacks that exploit these vulnerabilities.
\item \textit{Industry and Application} involves questions about applications, business sectors, and use cases of DLT systems. Relevant relations in this category are the applications of DLT systems in specific use cases, as well as the mapping of use cases to a business sector.
\item \textit{Standardization and Regulation} involves questions about standards, technical controls, standardization organizations, and relevant laws. From this category we derive relations regarding compliance to standards, as well as the technical control of specific DLT components.
\end{itemize}

\begin{table}[ht!]
    \caption{Competency Questions used to evaluate and validate the ontology.}
    \label{tab:questions}
    \centering
    \begin{tabularx}{\textwidth}{l X}
        \toprule
        ID \hspace{2pt}  & Competency Question \\
        \toprule
        \multicolumn{2}{l}{\bf Technology and Security} \\
        T1 & Which components are part of the distributed ledger system?\\
        T2 & Which technical threats have to be considered regarding the consensus algorithm of the system?\\
        T3 & What are known smart contract vulnerabilities?\\
        T4 & What are the data structures of the system?\\
        T5 & Which types of DLT attacks could be used against a system or its components?\\
        \midrule
        \multicolumn{2}{l}{\bf Industry and Application} \\
        I1 & Which use cases can be realised with distributed ledger technology?\\
        I2 & Which industries could use distributed ledger systems?\\
        I3 & Which distributed ledger systems are used for public transportation and smart cars?\\
        I4 & Which types of record keeping could be realized with distributed ledger technology?\\
        \midrule
        \multicolumn{2}{l}{\bf Standardization and Regulation} \\
        S1 & Which standardization organizations are active in regards to distributed ledger technology?\\
        S2 & Which normative references do exist for distributed ledger systems?\\
        S3 & Which industry standards do exist?\\
        S4 & What are relevant laws in regards to distributed ledger systems?\\
        S5 & What are organizational controls and mitigations for a distributed ledger system and/or component?\\
        S6 & Is there an industry initiative that directs and regulates the used distributed ledger system?\\
        \bottomrule
    \end{tabularx}
\end{table}

In Listing \ref{lst:sparql_t2}, \ref{lst:sparql_i5}, and \ref{lst:sparql_s1} we give three SPARQL translations of the competency questions. In \ref{lst:sparql_t2} we ask for existing technical threats that potentially threaten the consensus algorithm of a system; in \ref{lst:sparql_i5} we translate question I4, which asks for specific use cases of a DLT system; the query in Listing \ref{lst:sparql_s1} lists standardization organizations that actively publish DLT standards. 

\begin{figure}[htb]
\begin{lstlisting}[captionpos=b, caption=T2 - Which technical threats have to be considered regarding the consensus algorithm of the system?, label=lst:sparql_t2, basicstyle=\scriptsize\ttfamily,frame=single]
PREFIX : <https://w3id.org/DLTOntology#>

SELECT ?threat WHERE {
  ?threat :threatens [ a :ConsensusAlgorithm ]
}
\end{lstlisting}
\begin{lstlisting}[captionpos=b, caption=I4 - Which types of record keeping could be realized with distributed ledger technology?, label=lst:sparql_i5, basicstyle=\scriptsize\ttfamily,frame=single]
SELECT ?dltsystem ?usecase WHERE {
  ?dltsystem :isSpecializedFor ?usecase .
  ?usecase a :RecordKeeping .
}
\end{lstlisting}
\begin{lstlisting}[captionpos=b, caption=S1 - Which standardization organizations are active in regards to distributed ledger technology?, label=lst:sparql_s1, basicstyle=\scriptsize\ttfamily,frame=single]
SELECT ?stdorga ?standard WHERE {
  ?stdorga a :StandardizationOrganization ;
           :creates ?standard .
  ?dltsystem :compliantTo ?standard .
}
\end{lstlisting}
\end{figure}

The complete list of questions as SPARQL queries for all competency questions, including the respective classes and properties, can be found in the online documentation: \url{https://w3id.org/DLTOntology}.

\section{DLT Ontology and Knowledge Graph} \label{sec:Ontology}
Based on the collected information and competency questions we developed \textit{DLT Ontology}. The ontology covers the technical setup and components of a DLT system, security aspects, as well as its applications and use cases.
It covers 115 classes and 15 properties, and consists of a total of 571 triples.

\begin{figure}[ht]
    \centering
    \includegraphics[width=0.8\textwidth]{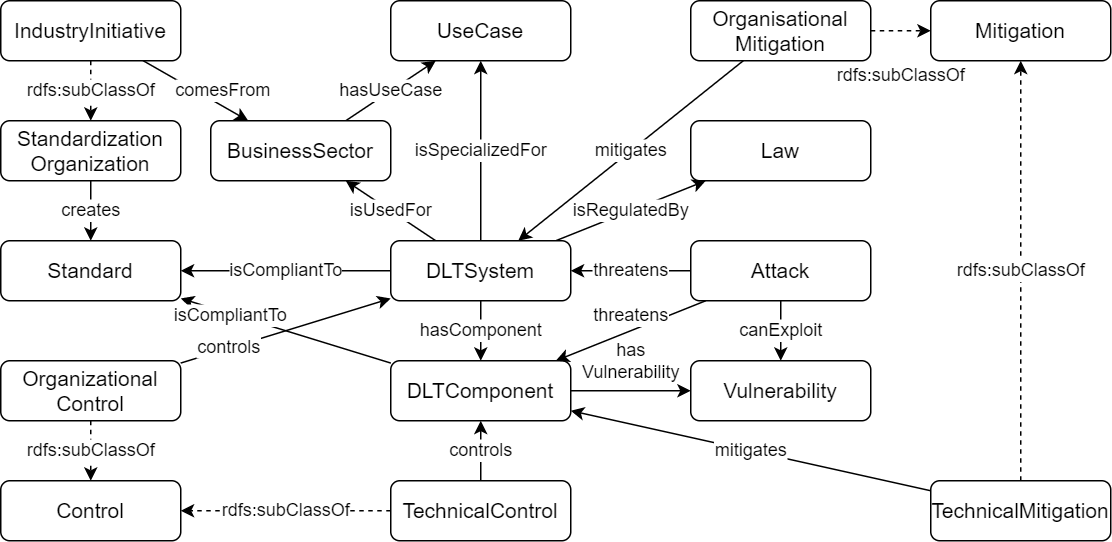}
    \caption{Overview of the core concepts of the DLT Ontology.}
    \label{fig:core_concepts}
\end{figure}

The DLT ontology provides classes and properties which are used to describe various aspects of a distributed ledger ecosystem. 
The core concepts are displayed in Figure~\ref{fig:core_concepts}: \textit{DLTSystem} and \textit{DLTComponent} describe the essential parts of a DLT system; \textit{Vulnerability} and \textit{Attack} connect potential security aspects of the system; \textit{UseCase}, and \textit{BusinessSector} link existing applications of the described systems.

The relations between the included classes are described in detail in Table \ref{tab:Properties}, where a property-centric view of the core concepts, including domains and ranges of the properties of the ontology is provided.

\begin{table}[p]
    \centering
    \caption{Main Properties of the DLT Ontology.}
    \label{tab:Properties}
    \begin{adjustbox}{width=1\textwidth}{}
        \begin{tabularx}{420pt}{l|X}
        \hline
            \bf Property & \bf Description \\
        \hline
        \hline
        \makecell[tl]{
            \textbf{canExploit} \\
            \textit{Domain:} Attack \\
            \textit{Range:} Vulnerability}
            & The \emph{canExploit} property links an attack on a distributed ledger system or component to the respective vulnerability, which could be used by an attacker to enable further malicious activity in an effort to manipulate or gain control of the system.\\
        \hline
        \makecell[tl]{
            \textbf{threatens} \\
            \textit{Domain:} Attack \\
            \textit{Range:} DLTSystem/Component}
             & This property indicated that there is an existing attack on a distributed ledger system or one of its components. Attacks can manipulate specific components and data, or generally weaken the structural integrity of the entire system.\\
        \hline
        \makecell[tl]{
            \textbf{hasVulnerability} \\
            \textit{Domain:} DLTComponent \\
            \textit{Range:} Vulnerability}
             & This property indicates that one or multiple of the components used in a distributed ledger system carry with them one or an array of vulnerabilities. Having an overview of existing vulnerabilities is crucial for securing the system.\\
        \hline
        \makecell[tl]{
            \textbf{mitigates} \\
            \textit{Domain:} Technical/Org.Mitigation \\
            \textit{Range:} DLTSystem/Component}
             & Once disaster strikes, it is important to have a series of mitigations in place, to reduce the impact of an attack, malicious event or generally in case of the system breaking down. This property links a technical or organizational mitigation measure to a distributed ledger system or specific component.\\
        \hline
        \hline
        \makecell[tl]{
            \textbf{hasComponent} \\
            \textit{Domain:} DLTSystem \\
            \textit{Range:} DLTComponent}
             & It links an individual component to the DLT system. It is used to show what a system is comprised of and to get a better understanding of the elements and building blocks. Such an overview allows for an analysis or the implementation of measures for a single component specifically and to tailor solutions to certain requirements.\\
        \hline
        \makecell[tl]{
            \textbf{isCompliantTo} \\
            \textit{Domain:} DLTSystem/Component \\
            \textit{Range:} Standard}
             & When it comes to using distributed ledger systems in an enterprise setting, national and/or international standardization can become a necessity. This property expresses compliance to existing standardization or normative references on a system or component level.\\
        \hline
        \makecell[tl]{
            \textbf{controls} \\
            \textit{Domain:} Technical/Org.Control \\
            \textit{Range:} DLTComponent/System}
             & Implemented controls can secure components and the entire system itself. This property links a technical or organizational control measure to a distributed ledger system or individual component.\\
        \hline
        \makecell[tl]{
            \textbf{createsStandard} \\
            \textit{Domain:} Stand.Organization \\
            \textit{Range:} Standard}
             & Behind every normative reference material and standardization there is an organization with the needed authority to publish such material, which can be either a national institute or an international organization. This property links a standard to its organization.\\
        \hline
        \makecell[tl]{
            \textbf{isRegulatedBy} \\
            \textit{Domain:} DLTSystem \\
            \textit{Range:} Law}
             & As with regular ICT-systems, a distributed ledger system will have to comply with the legal status of the respective country it is used in by any organization. This property connects existing legislation for ICT and DLT systems.\\
        \hline
        \hline
        \makecell[tl]{
            \textbf{isSpecializedFor} \\
            \textit{Domain:} DLTSystem \\
            \textit{Range:} UseCase}
             & Not all implementations of distributed ledger systems are the same or similar in their nature and complexion. This property links existing DLT systems which are tailored for the delivery of a specific use case.\\
        \hline
        \makecell[tl]{
            \textbf{hasUseCase} \\
            \textit{Domain:} BusinessSector \\
            \textit{Range:} UseCase}
             & This property is used to show which use cases could be used in a selected industry. Not all forms of usage of distributed ledger systems are fit for all types of business.\\
        \hline
        \makecell[tl]{
            \textbf{isUsedFor} \\
            \textit{Domain:} DLTSystem \\
            \textit{Range:} BusinessSector}
             & There are certain distributed ledger systems that are specialized for the use in one specific field or industry. This property is used to highlight this exact relation.\\
        \hline
        \makecell[tl]{
            \textbf{comesFrom} \\
            \textit{Domain:} IndustryInitiative \\
            \textit{Range:} BusinessSector}
             & It is entirely possible for several organizations of an industry to form an initiative to steer and develop the future of distributed ledger technology within that industry in a common effort. This property shows the origins of an industry initiative with regards to the industry of involved parties.\\
        \hline
        \makecell[tl]{
            \textbf{hasBusinessSector} \\
            \textit{Domain:} UseCase \\
            \textit{Range:} BusinessSector}
             & Distributed ledger technology can be used to realize a variety of different use cases. However, not all of them are a fit for any field or industry. This property highlights the connection between a use case and the specific business sector/industry it is used in.\\
        \hline
        \hline
        \end{tabularx}
    \end{adjustbox}
\end{table}


\subsection{Knowledge Graph\label{ssec:kgDetails}} 

The knowledge graph uses the above introduced DLT Ontology, and is based on the information collection described in \Cref{ssec:informationcollection}. It consists of three parts: (i) standards and legal authorities \cite{fi12120222}, (ii) technical details, vulnerabilities and security aspects \cite{jisis20-10-3-06}, and (iii) use cases and business sectors \cite{itu2019itu}.
Additionally, we reviewed white papers for further information about the technical collocation of DLT systems \cite{hawig2019designing}.\footnote{\url{https://www.r3.com/reports/corda-technical-whitepaper/}, \url{https://ethereum.org/en/whitepaper/}}

In total, the knowledge graph consists of 746 triples; \Cref{tab:kgdata} lists the number of entities for the core classes.

\begin{table}[ht]
    \caption{Number of triples and entities in the knowledge graph.}
    \label{tab:kgdata}
    \centering
    \begin{tabularx}{\textwidth}{c | c c c c c c c}
        \toprule
        \texttt{Triples} & \texttt{Standard} & \texttt{Std.Orga.} & \texttt{DLTComponent} & \texttt{UseCase} & \texttt{BusinessSector} & \texttt{Vuln.} & \texttt{Attack}  \\
        \midrule
        746 & 18 & 8 & 9 & 55 & 9 & 7 & 11 \\
        \bottomrule
    \end{tabularx}
\end{table}

\subsection{Availability of the Resources\label{sec:avail}}

\textbf{DLT Ontology and Knowledge Graph.} The knowledge graph and respective documentation are available online at \url{https://w3id.org/DLTOntology} under the CC-by-4.0 license; a DOI is provided via the Zenodo repository.\footnote{\url{https://doi.org/10.5281/zenodo.6497619}} The online documentation consists of a comprehensive overview, examples of how to use the ontology, and SPARQL queries for the respective competency questions.

\noindent\textbf{SPARQL Interface.} We provide a SPARQL query interface at \url{https://w3id.org/DLTOntology} which allows to execute the competency questions provided in \Cref{sec:evaluation}. The interface is based on the Comunica SPARQL Widget \cite{taelman_iswc_resources_comunica_2018}. 


\section{Related Work} \label{sec:relatedWork}

\paragraph{Glossaries and Vocabularies for Distributed Ledgers.}
There have been made various attempts and proposals on the creation of terminologies, glossaries and vocabularies regarding blockchains and distributed ledgers. Most importantly, there is the ISO working group TC 307 \cite{ISOvocabulary}, the ITU-T \emph{Focus Group on Application of Distributed Ledger Technology} \cite{ituTerms}, and the German Institute for Standardization (DIN) \cite{dinTerminology}. On top of that there is a multitude of private companies and blockchain enthusiasts that offer their own glossaries for blockchains and distributed ledgers like Blockchainhub Berlin\footnote{\url{https://blockchainhub.net/blockchain-glossary/}}, 101 Blockchains\footnote{\url{https://101blockchains.com/blockchain-definitions/}}, and ConsenSys \cite{ConsensysGlossary}. 


\paragraph{Ontologies for Distributed Ledgers.}
EthOn (the Ethereum Ontology) \cite{EthOn} is an approach to model the major concepts of the Ethereum blockchain, e.g., Blocks, Accounts, Transactions, etc. The ontology itself is based on OWL \cite{OWL} and there exist extensions that can be used to describe ERC20 compliant tokens\footnote{\url{https://github.com/ConsenSys/EthOn/tree/master/ERC20}} and  smart contracts.\footnote{\url{https://github.com/ConsenSys/EthOn/tree/master/Contracts}}
While EthOn focuses on modelling the technical details of a particular blockchain, we propose an approach that allows to model DLT systems, their specific use cases, and their potential security threats, independent of the concrete implementation.


Also related to our efforts is the BLONDiE ontology (Blockchain Ontology with Dynamic Extensibility) \cite{hector2020blondie}. 
BLONDiE provides classes and properties to describe the structure and related information of the three
most prominent blockchain projects -- Bitcoin, Ethereum and Hyperledger. However, while our goal is to cover the overall ecosystem and landscape of existing DLT technologies, BLONDiE describes concepts that specifically relate to the implementation of the three technologies (e.g., ``Ethereum Payload'', and ``Hyperledger Transaction'').


\section{Conclusion} \label{sec:conclusion}
Although distributed ledger technology has seen an enormous increase in usage and spread over the last decade, there are still many misconceptions and misunderstandings revolving around it. First and foremost is the equalization of distributed ledger technology with blockchains, where the latter is simply one of many manifestations of the former. Furthermore, there is a lack of holistic vocabularies and conceptualization of the technology as a whole, rather than just one of its manifestations or even a single product. To tackle and overcome present issues, we have contributed the following:
\begin{itemize}
    \item We developed an ontology for distributed ledgers. This ontology considers various aspects of DLT systems, including threats, vulnerabilities, and the legal situation and standardization of the technology; an overview of the market situation and real-world applications is also included. 
    \item We have demonstrated the use of the ontology by building a knowledge graph containing entities and relations for all the core classes of the ontology, i.e. (i) standards and legal authorities, (ii) components and technical details of DLT systems, as well as corresponding vulnerabilities and (iii) exemplary use cases.
    \item We have validated the ontology based on a pre-defined set of competency questions. On top of that, a set of SPARQL queries is provided to evaluate the competency questions and thus the capabilities of the knowledge graph.
    \item The ontology and the knowledge graph are available for public use under an open license; the documentation and downloads can be found at \url{https://w3id.org/DLTOntology}.
\end{itemize}


One limitation of our work in this regard is the lack of existing data that can be integrated in a knowledge graph: while it was possible to gather information on the technical components of DLT systems, existing standardization documents, and potential use cases in different industries, this requires intensive research, extensive manual work, and expert knowledge in the domain.

\begin{acks}
This research was funded by the Josef Ressel Center for Blockchain Technologies \& Security Management (BLOCKCHAINS). Sebastian Neumaier received funding through the Austrian Research Promotion Agency (FFG) Bridge project 880592 ``SecDM -- Sichere und vertrauens-würdige on-premise data markets''. The financial support by the FFG and the Christian Doppler Research Association is gratefully acknowledged.
\end{acks}

\bibliographystyle{ios1}
\bibliography{references.bib}

\end{document}